\documentclass{article}

    \PassOptionsToPackage{numbers, compress}{natbib}

\usepackage[preprint]{neurips_2023}
\usepackage{amsmath}
\usepackage{multirow}
\usepackage{graphicx}
\usepackage[labelformat=simple,labelsep=colon]{subcaption}

\usepackage[utf8]{inputenc} 
\usepackage[T1]{fontenc}    
\usepackage{hyperref}       
\usepackage{url}            
\usepackage{booktabs}       
\usepackage{amsfonts}       
\usepackage{nicefrac}       
\usepackage{microtype}      
\usepackage{xcolor}         

\title{Automatic Creative Selection with Cross-Modal Matching}

  \author{%
  Alex Kim\thanks{work completed as an intern at Apple}\\
  University of Southern California\\
  Los Angeles, CA \\
  \texttt{ajkim@usc.edu} \\
  \And
  Jia Huang\thanks{corresponding author}\\
  Apple \\
  Cupertino, CA \\
  \texttt{jhuang57@apple.com} \\
  \And
  Rob Monarch \\
  Apple \\
  Cupertino, CA \\
  \texttt{rmonarch@apple.com} \\
  \And
  Jerry Kwac \\
  Apple \\
  Cupertino, CA \\
  \texttt{jkwac@apple.com} \\
  \And
  Anikesh Kamath \\
  Apple \\
  Cupertino, CA \\
  \texttt{anikesh@apple.com} \\
  \And
  Parmeshwar Khurd \\
  Apple \\
  Cupertino, CA \\
  \texttt{pkhurd@apple.com} \\
  \AND
  Kailash Thiyagarajan \\
  Apple \\
  Cupertino, CA \\
  \texttt{k\_thiyagarajan@apple.com} \\
  \And
  Goodman Gu \\
  Apple \\
  Cupertino, CA \\
  \texttt{xiaoyuan\_gu@apple.com} 
}

\begin{document}

\maketitle

\begin{abstract}
We present a novel approach for matching images to text in the specific context of matching an application image to the search phrases that someone might use to discover that application. We share a new fine-tuning approach for a pre-trained cross-modal model, tuned to search-text and application-image data. We evaluate matching images to search phrases in two ways: the application developers' intuitions about which search phrases are the most relevant to their application images; and the intuitions of professional human annotators about which search phrases are the most relevant to a given application. Our approach achieves 0.96 and 0.95 AUC for these two ground truth datasets, which outperforms current state-of-the-art models by 8\%-17\%.  
\end{abstract}

\section{Introduction}

Application (App) developers promote their Apps by creating multiple custom pages, each with different images. By using different images for different pages, the App developers can engage people with diverse interests to use their Apps. In some cases, developers also advertise their Apps and in these cases they suggest which search phrases that they think are the most relevant for their Apps.

This give us two different but equally meaningful but subjective perspectives on the relationship between Apps and search phrases: the intuition of the people who created each App and the intuition of people in general. Therefore, understanding the relationship between images and search terms, and providing a model to recommend creative images to developers that best match with search phrases, is an important task for supporting App developers. Recent state-of-the-art approaches have addressed this problem using a cross-modal Transformer, or BERT architecture
(\cite{chen2020uniter}\cite{fei2021cross}\cite{gan2020large}\cite{hong2021gilbert}\cite{li2020unicoder}\cite{li2020oscar}\cite{radford2021learning}\cite{yu2021heterogeneous}).

The existing approaches work well for image captioning and QA tasks, but in the search domain creative images often do not have a description, and search phrases contain much shorter text, making description-based matching infeasible. We customizes a cross-modal BERT framework by fine-tuning a pre-trained cross-modal model on an in-house training dataset of (search phrase, ad image, label), and evaluate the model on our test set for the (image, search phrase) matching task and compare with baselines including CLIP \cite{radford2021learning}.

\section{Method}
For each App ${c}$, a developer promoting that App can define a set of search phrases ${K}$ relevant to their App. Our approach provides a way to automatically select an image ${m}$ from a candidate image pool ${M}$ that best matches a given search phrase ${k}$. Given a pair of (search phrase ${k}$, image ${m}$), ${c}$, we model relevance ${R(k, m)}$ prediction as binary classification. 

Our model uses pre-trained cross-modal image-text matching architecture, as in LXMERT \cite{tan2019lxmert}, where input text is split by a WordPiece tokenizer \cite{wu2016google}, and embedded to word embeddings, and input image objects are detected by Faster R-CNN \cite{ren2015faster} and embedded to object-level image embeddings. A language encoder and object-relationship encoder are applied to word embeddings and detected objects respectively. Finally, the relevance score ${R(k, m)}$ is predicted with a Transformer-based cross-modal encoder. Extracted features from each image are in the format of ([36, 4], [36, 2048]) representing bounding box coordinates of up to 36 detected objects and 2048-dim feature vector of those objects. We then fine-tune the pre-trained model on our training dataset. Our fine-tuning process is defined as a sequential model built on top of a LXMERT Encoder, which consists of a linear layer, a GELU activation function \cite{hendrycks2016gaussian}, a layer normalization \cite{ba2016layer}, another linear layer, and a sigmoid activation function as the last layer to output binary classification predictions. 


\section{Results}

\begin{table}
  \caption{Model performance on two evaluation datasets}
  \label{table: 2}
   \centering
  \begin{tabular}{ccc|cc}
    \toprule
    &\multicolumn{2}{c}{Developer intuitions}&\multicolumn{2}{|c}{Professional annotator intuitions}\\
    &AUC&F1&AUC&F1\\
    \midrule
    Zero-shot CLIP & 0.62 & 0.70  & 0.74 & 0.68 \\
    Fine-tuned CLIP & 0.84 & 0.81 & 0.81 & 0.75\\
    XLM-R + ResNet & 0.89 & 0.76 & 0.82 & 0.73  \\
    XLM-R + CLIP img & - & - & 0.83 & 0.75 \\
    Our Approach & \textbf {0.96} & \textbf {0.89} & \textbf {0.95} & \textbf{0.87}\\
  \bottomrule
\end{tabular}
\end{table}

We compare our method to four baselines: (1) XLM-R + ResNet (2) XLM-R + CLIP img (3) Zero-shot CLIP and (4) Fine-tuned CLIP as shown in Table \ref{table: 2}. One possible reason for the performance lift lies in where the fusion occurs. While both baselines use early-fusion, where text and image features are concatenated as input sequence, our approach uses mid-fusion where independent transformers are applied to the textual and the visual modality, then a cross-modal encoder is applied, showing the effectiveness of the cross-modal encoder in identifying the relationship between modalities. 

CLIP uses a contrastive loss assuming a 1:1 mapping between text and image, so our lift here is likely due to our 1:n mapping. CLIP also follows ``shallow-interaction design'' \cite{shen2021much}, with no cross-modal encoder, so the the relationship between text and images might be too shallow for our use case.

\section{Conclusion}
We demonstrate that a cross-modal model framework significantly improves the prediction accuracy for image-text matching in the domain of applications and search phrases. Our work opens doors to automatic image selection and for self-serve capabilities for developers that recommendation which image is the best to promote their applications.

{
\small
\bibliographystyle{ACM-Reference-Format}
\bibliography{automatic_creative_selection_baylearn23}
}






\end{document}